# Comparison-Based Learning with Rank Nets

**Amin Karbasi** AMIN.KARBASI@EPFL.CH
EPFL, Lausanne, Switzerland

**Stratis Ioannidis** STRATIS.IOANNIDIS@TECHNICOLOR.COM
Technicolor, Palo Alto, USA

**Laurent Massoulié** LAURENT.MASSOULIE@TECHNICOLOR.COM
Technicolor, Paris, France

## Abstract

We consider the problem of search through comparisons, where a user is presented with two candidate objects and reveals which is closer to her intended target. We study adaptive strategies for finding the target, that require knowledge of rank relationships but not actual distances between objects. We propose a new strategy based on rank nets, and show that for target distributions with a bounded *doubling constant*, it finds the target in a number of comparisons close to the entropy of the target distribution and, hence, of the optimum. We extend these results to the case of noisy oracles, and compare this strategy to prior art over multiple datasets.

## 1. Introduction

In search through comparisons, a user locates a target object in a database as follows. At each step, the database presents two objects to the user, who then selects among the pair the object closest to the target that she has in mind. This process continues until, based on the user's answers, the database can uniquely identify the target she has in mind.

This kind of interactive navigation, also known as exploratory search, has numerous real-life applications (Marchionini, 2006; Ruthven, 2008), such as navigation in a database of pictures of people photographed in an uncontrolled environment (Tschopp et al., 2011). Automated methods may fail to extract meaningful features from such photos. Even if this were possible, in many practical cases, images with similar low-level descriptors may have very different semantic content, and thus be perceived differently by users (Smeulders et al., 2000; Lew et al., 2006). On the other hand, a human can easily sort images of people w.r.t. their similarity to a given person, and her answers can be used to rank images in the database in terms of this similarity.

Formally, the human user's feedback can be modelled as a "comparison oracle" (Goyal et al., 2008). Assuming a database $\mathcal{N}$ endowed with a distance metric $d$, capturing the "distance" or "dissimilarity" between different objects, a comparison oracle answers questions of the kind: "Between two objects $x$ and $y$ in $\mathcal{N}$, which one is closest to $t$ under the metric $d$?".

In this paper, we study algorithms for identifying an unknown target with as few queries to such an oracle as possible. Most importantly, the algorithms we consider do not rely on a priori knowledge of the distance between objects: they *cannot* access an embedding of $\mathcal{N}$ in a metric space, nor can they compute the distance between two objects. Decisions on which queries to submit to the oracle depend only on (a) *ranking relationships* between objects, which can indeed be obtained through a comparison oracle and (b) the *prior distribution* $\mu$ from which the target is sampled.

As discussed in Section 3.3, content search through comparisons can be framed as an active learning problem. A well-known active learning algorithm is the *Generalized Binary Search* (GBS) or *splitting* algorithm (Dasgupta, 2005). Using GBS to submit queries to the oracle locates the target in $OPT \cdot \big(H_{\max}(\mu)+1\big)$ queries, where $H_{\max}(\mu) = \max_{x\in\mathsf{supp}(\mu)} \log\frac{1}{\mu(x)}$ and $OPT$ is the number of queries submitted by an optimal algorithm. In practice, GBS performs very well in terms of query complexity, suggesting that this bound can be tightened. However, the computational complexity of GBS is $\Theta(n^3)$ for $n = |\mathcal{N}|$, which makes it intractable for most large databases.

Recently, Karbasi *et al.* (2011) proposed an algorithm that determines the target in $O(c^3 H(\mu) H_{\max}(\mu))$ queries, whose computational complexity is $O(1)$ per query. Here, $H(\mu)$ is the entropy of the prior $\mu$ and $c$, defined formally in Section 3.2, is the *doubling constant* of the prior $\mu$. It captures the dimension of the database, as determined by the underlying distance $d$ (Clarkson, 2006). Karbasi *et al.* also show that $OPT = \Omega(cH(\mu))$, indicating that their algorithm is within a $c^2 H_{\max}(\mu)$ factor from the optimal.

We make the following contributions: First, we propose a new adaptive algorithm, RankNetSearch, locating the target with $O(c^6 H(\mu))$ queries to the oracle, in expectation. Our algorithm therefore improves on GBS and Karbasi *et al.* by removing the term $H_{\max}$—which can be quite large in practice—at a cost of a higher exponent in the dependence on the constant $c$. Its computational complexity is $O\big(n(\log n + c^6)\log c\big)$ per query, which is manageable compared to GBS; moreover, this cost can be reduced to $O(1)$ by precomputing an additional data structure.

Second, we extend RankNetSearch to the case of a *faulty oracle* that lies with a probability $\epsilon > 0$, and show that it locates the target *w.h.p.* at an expected query cost $O(\sum_{x\in\mathcal{N}} \mu(x) \log \frac{1}{\mu(x)} \log\log(\mu(x))$ and, thereby, close to $H(\mu)$.

Third, we evaluate RankNetSearch and prior art algorithms over several datasets. We observe that RankNetSearch establishes a desirable trade-off between query and computational complexity.

The remainder of this paper is organized as follows. We overview related work in Section 2, and discuss definitions and preliminaries in Section 3. Our algorithm and the analysis of its complexity are presented in Section 4, and its robustness to noise in Section 5. Section 6 includes our numerical evaluations.

# 2. Related Work

Search through comparisons was first introduced by Goyal *et al.* (2008), and further explored by Lifshitz and Zhang (2009) and Tschopp *et al.* (2011). The above works study the problem in terms of worst-case (prior-free) bounds, so our work departs in introducing a prior $\mu$ and studying query complexity in expectation. All these works introduce a *disorder constant*, that plays the same role as the quantity $c$ in our setup. Lifshitz and Zhang also employ hierarchical data structures similar to the rank-nets we study here. Our upper bound coincides with theirs when $\mu$ is uniform over $\mathcal{N}$ and can thus be seen as an extension to the more general Bayesian setting under prior $\mu$.

Cover trees based on nets have been extensively studied in the context of nearest neighbour search (Clarkson, 1999; Beygelzimer et al., 2006). These works too focus on worst-case bounds and, crucially, assume full access to the underlying distance metric $d$. Our approach thus differs in both of these respects. In earlier work, Fredman (1976) and others have considered decision trees for determining a complete ordering of objects rather than just the first one in the list.

To the best of our knowledge, the work closest to our is (Karbasi et al., 2011), which was the first to study search through comparisons in a Bayesian setting. Our work improves their bound by the factor of $H_{\max}$ and establishes the connection to active learning and GBS.

# 3. Definitions and Preliminaries

## 3.1. Search through Comparisons

Consider a large finite set of objects $\mathcal{N}$ of size $n = |\mathcal{N}|$, endowed with a distance metric $d$, capturing the "dissimilarity" between objects. A user selects a target $t \in \mathcal{N}$ from a prior distribution $\mu$; our goal will be to design an interactive algorithm that queries the user with the purpose of discovering $t$.

**Comparison Oracle.** Though we assume that the metric $d$ exists, our view of distances is constrained to only observing order relationships. More precisely, we only have access to information that can be obtained through a *comparison oracle* (Goyal et al., 2008). Given object $z$, a comparison oracle $\mathcal{O}_z$ receives as a query an ordered pair $(x, y) \in \mathcal{N}^2$ and answers the question "is $z$ strictly closer to $x$ than to $y$?", *i.e.*,

$$\mathcal{O}_z(x,y) = \begin{cases} +1 & \text{if } d(x,z) < d(y,z), \\ -1 & \text{if } d(x,z) \geq d(y,z) \end{cases} \quad (1)$$

Note that a tie $d(x,z) = d(y,z)$ is revealed by two calls $\mathcal{O}_z(x,y)$ and $\mathcal{O}_z(y,x)$. Our algorithm for determining the unknown target $t$ can submit queries to a comparison oracle $\mathcal{O}_t$—namely, the user. We thus assume, effectively, that the user can order objects w.r.t. their distance from $t$, but does not need to disclose (or even know) the exact values of these distances. We will first assume that the oracle always gives correct answers; in Section 5, we relax this assumption by considering a *faulty oracle* that lies with probability $\epsilon < 0.5$.

**Prior Knowledge and Performance Metrics.** The algorithms we study rely only on a priori knowledge of (a) the distribution $\mu$ and (b) the values of the mapping $\mathcal{O}_z : \mathcal{N}^2 \to \{-1, +1\}$, for every $z \in \mathcal{N}$. This is in line with our assumption that, although the distance metric $d$ exists, it cannot be directly observed.

Our focus is on *adaptive* algorithms, whose decision on which query in $\mathcal{N}^2$ to submit next are determined by the oracle's previous answers.

The prior $\mu$ can be estimated empirically as the frequency with which objects have been targets in the past. The order relationships can be computed off-line by submitting $\Theta(n^2 \log n)$ queries to a comparison oracle, and requiring $\Theta(n^2)$ space: for each possible target $z \in \mathcal{N}$, objects in $\mathcal{N}$ can be sorted w.r.t. their distance from $z$ with $\Theta(n \log n)$ queries to $\mathcal{O}_z$. We store the result of this sorting in (a) a linked list, whose elements are sets of objects at equal distance from $z$, and (b) a hash-map, that associates every element $y$ with its rank in the sorted list. Note that $\mathcal{O}_z(x, y)$ can thus be retrieved in $O(1)$ time by comparing the relative ranks of $x$ and $y$ with respect to their distance from $z$.

We measure the performance of an algorithm through two metrics. The first is the *query complexity*, determined by the expected number of queries the algorithm needs to submit to the oracle to determine the target. The second is the *computational complexity*, determined by the time-complexity of determining the query to submit to the oracle at each step.

### 3.2. A Lower Bound

Recall that the *entropy* of $\mu$ is defined as $H(\mu) = \sum_{x \in \mathsf{supp}(\mu)} \mu(x) \log \frac{1}{\mu(x)}$ where $\mathsf{supp}(\mu)$ is the support of $\mu$. Given an object $x \in \mathcal{N}$, let $B_x(r) = \{y \in \mathcal{N} : d(x, y) \le r\}$ be the closed ball of radius $r \ge 0$ around $x$. Given a set $A \subseteq \mathcal{N}$ let $\mu(A) = \sum_{x \in A} \mu(x)$. The *doubling constant*[1] $c(\mu)$ of a distribution $\mu$ is the minimum $c > 0$ for which $\mu(B_x(2R)) \le c \cdot \mu(B_x(R))$, for any $x \in \mathsf{supp}(\mu)$ and any $R \ge 0$.

The doubling constant has a natural connection to the underlying dimension of the dataset (Clarkson, 2006; Karbasi et al., 2011), as determined by the distance $d$. Both the entropy and the doubling constant are also inherently connected to content search through comparisons. Karbasi *et al.* show that any adaptive mechanism for locating a target $t$ must submit at least $\Omega\big(c(\mu)H(\mu)\big)$ queries to the oracle $\mathcal{O}_t$, in expectation. Moreover, they provide an algorithm for determining the target in $O\big(c^3 H(\mu) H_{\max}(\mu)\big)$ queries, where $H_{\max}(\mu) = \max_{x \in \mathsf{supp}(\mu)} \log \frac{1}{\mu(x)}$.

### 3.3. Active Learning

Search through comparisons can be seen as a special case of *active learning* (Dasgupta, 2005; Nowak, 2012). In active learning, a *hypothesis space* $\mathcal{H}$ is a set of binary valued functions defined over a finite set $\mathcal{Q}$, called the *query space*. Each hypothesis $h \in \mathcal{H}$ generates a label from $\{-1, +1\}$ for every query $q \in \mathcal{Q}$. A target hypothesis $h^*$ is sampled from $\mathcal{H}$ according to some prior $\mu$; asking a query $q$ amounts to revealing the value of $h^*(q)$, thereby restricting the possible candidate hypotheses. The goal is to determine $h^*$ in an adaptive fashion, by asking as few queries as possible.

In our setting, the hypothesis space $\mathcal{H}$ is the set $\mathcal{N}$, and the query space $\mathcal{Q}$ is the set of ordered pairs $\mathcal{N}^2$. The target hypothesis sampled from $\mu$ is the unknown target $t$. Each hypothesis/object $z \in \mathcal{N}$ is uniquely[2] identified by the mapping $\mathcal{O}_z : \mathcal{N}^2 \to \{-1, +1\}$, which we have assumed to be a priori known.

**Generalized Binary Search** A well-known algorithm for determining the true hypothesis in the general active-learning setting is the so-called *generalized binary search* (GBS) or *splitting* algorithm (Dasgupta, 2005; Nowak, 2012). Define the *version space* $V \subseteq \mathcal{H}$ to be the set of possible hypotheses that are consistent with the query answers observed so far. At each step, GBS selects the query $q \in \mathcal{Q}$ that minimizes $|\sum_{h \in V} \mu(h) h(q)|$. Put differently, GBS selects the query that separates the current version space into two sets of roughly equal probability mass; this leads, in expectation, to the largest reduction in the mass of the version space as possible, so GBS can be seen as a greedy query selection policy.

A bound on the query complexity of GBS originally obtained by Dasgupta (2005) and recently tightened (w.r.t. constants) by Golovin and Krause (2010) is given by the following theorem:

**Theorem 1.** *GBS makes at most* $OPT \cdot \big(H_{\max}(\mu) + 1\big)$ *queries in expectation to identify hypothesis* $h^* \in \mathcal{N}$*, were* $OPT$ *is the minimum expected number of queries made by any adaptive policy.*

**GBS in Search through Comparisons.** In our setting, the version space $V$ comprises all possible objects in $z \in \mathcal{N}$ that are consistent with oracle answers given so far. In other words, $z \in V$ iff $\mathcal{O}_z(x, y) = \mathcal{O}_t(x, y)$ for all queries $(x, y)$ submitted to the oracle. Selecting the next query therefore amounts to finding the pair $(x, y) \in \mathcal{N}^2$ that minimizes

$$f(x, y) = \big| \sum_{z \in V} \mu(z) \mathcal{O}_z(x, y) \big|. \quad (2)$$

As the simulations in Section 6 show, the query complexity of GBS is excellent in practice. This suggests

[1] $c$ relates to the *doubling dimension* $\delta$ through $c = 2^\delta$.

[2] Note that, for any two objects/hypotheses $z, z' \in \mathcal{N}$, there exists at least one query in $\mathcal{N}^2$ that differentiates them, namely $(z', z)$.

**Algorithm 1** RankNetSearch($\mathcal{O}_t$)

**Input**: Oracle $\mathcal{O}_t$
**Output**: Target $t$
Let $E \leftarrow \mathcal{N}$; select arbitrary $x \in E$
**repeat**
  ( $\mathcal{R}$, $\{B_y(r_y)\}_{y\in\mathcal{R}}$ )$\leftarrow$RankNet($x$,$E$)
  Find $y^*$, the object in $\mathcal{R}$ closest to $t$, using $\mathcal{O}_t$.
  Let $E \leftarrow B_{y^*}(r_{y^*})$ and $x \leftarrow y^*$;
**until** $E$ is a singleton
**return** $y$

**Algorithm 2** RankNet($x$,$E$)

**Input**: Root object $x$, Ball $E = B_x(R)$
**Output**: $\rho$-rank net $\mathcal{R}$, Voronoi balls $\{B_y(r_y)\}_{y\in\mathcal{R}}$
$\rho \leftarrow 1$
**repeat**
  $\rho \leftarrow \rho/2$; construct a $\rho$-net $\mathcal{R}$ of $E$
  $\forall\ y \in \mathcal{R}$, construct ball $B_y(r_y)$
  Let $\mathcal{I} \leftarrow \{y \in E : |B_y(r_y)| > 1\}$
**until** $\mathcal{I} = \emptyset$ **or** $\max_{y\in\mathcal{I}} \mu(B_y(r_y)) \leq 0.5\mu(E)$
**return** ($\mathcal{R}$,$\{B_y(r_y)\}_{y\in\mathcal{R}}$)

that the bound of Theorem 1 could be improved in the specific context of search through comparisons.

Nevertheless, the computational complexity of GBS is $\Theta(n^2|V|)$ operations per query, as it requires minimizing $f(x, y)$ over all pairs in $\mathcal{N}^2$. For large sets $\mathcal{N}$, this can be truly prohibitive. This motivates us to propose a new algorithm, RankNetSearch, whose computational complexity is almost linear and its query complexity is within a $O(c^5(\mu))$ factor from the optimal.

## 4. An Efficient Adaptive Algorithm

Our algorithm is inspired by $\epsilon$-nets, a structure introduced by Clarkson (1999; 2006) in the context of Nearest Neighbor Search (NNS). The main challenge that we face is that, contrary to standard NNS, we have *no access to the underlying distance metric.* In addition, the query complexity bounds on $\epsilon$-nets are worst-case (*i.e.*, prior-free); our construction takes the prior $\mu$ into account to provide bounds in expectation.

### 4.1. Rank Nets

To address the above issues, we introduce the notion of *rank nets*, which will play the role of $\epsilon$-nets in our setting. For some $x \in \mathcal{N}$, consider the ball $E = B_x(R) \subseteq \mathcal{N}$. For any $y \in E$, we define

$$d_y(\rho, E) = \inf\{r : \mu(B_y(r)) \geq \rho\mu(E)\} \tag{3}$$

to be the radius of the smallest ball around $y$ that maintains a mass above $\rho\mu(E)$. Using this definition[3], we define a $\rho$-rank net as follows.

**Definition 1.** *For some $\rho < 1$, a $\rho$-rank net of $E = B_x(r) \subseteq \mathcal{N}$ is a maximal[4] set of objects $\mathcal{R} \subset E$ such that for any two distinct $y, y' \in \mathcal{R}$*

$$d(y, y') > \min\{d_y(\rho, E), d_{y'}(\rho, E)\}. \tag{4}$$

For any $y \in \mathcal{R}$, consider the Voronoi cell $V_y = \{z \in E : d(y, z) \leq d(y', z), \forall y' \in \mathcal{R}, y' \neq y\}$. We also define the radius $r_y$ of the Voronoi cell $V_y$ as $r_y = \inf\{r : V_y \subseteq B_y(r)\}$. Critically, a rank net and the Voronoi tessellation it defines *can both be computed using only ordering information*:

**Lemma 1.** *A $\rho$-rank net $\mathcal{R}$ of $E$ can be constructed in $O(|E|(\log|E| + |\mathcal{R}|))$ steps, and the balls $B_y(r_y) \subset E$ circumscribing the Voronoi cells around $\mathcal{R}$ can be constructed in $O(|E||\mathcal{R}|)$ steps using only (a) $\mu$ and (b) the mappings $\mathcal{O}_z : \mathcal{N}^2 \to \{-1, +1\}$ for every $z \in E$.*

The proof is in Appendix A. Armed with this result, we turn our attention to how the selection of $\rho$ affects the size of the net as well as the mass of the Voronoi balls around it. Our next lemma, whose proof is in Appendix B, bounds $|\mathcal{R}|$.

**Lemma 2.** *The size of the net $\mathcal{R}$ is at most $c^3/\rho$.*

Finally, our last lemma determines the mass of the Voronoi balls in the net.

**Lemma 3.** *If $r_y > 0$ then $\mu(B_y(r_y)) \leq c^3\rho\mu(E)$.*

The proof is in Appendix C. Note that Lemma 3 *does not* bound the mass of Voronoi balls of radius zero.

### 4.2. Rank Net Data Structure and Algorithm

Rank nets can be used to identify a target $t$ using a comparison oracle $\mathcal{O}_t$ as described in Algorithm 1. Initially, a net $\mathcal{R}$ covering $\mathcal{N}$ is constructed; nodes $y \in \mathcal{R}$ are compared w.r.t. their distance from $t$, and the closest to the target is determined, say $y^*$. Note that this requires submitting $|\mathcal{R}| - 1$ queries to the oracle. The version space $V$ (the set of possible hypotheses) is thus the Voronoi cell $V_{y^*}$, and is a subset of the ball $B_{y^*}(r_{y^*})$. The algorithm then proceeds by limiting the search to $B_{y^*}(r_{y^*})$ and repeating the above process. Note that, at all times, the version space is included in the current ball to be covered by a net. The process terminates when this ball becomes a singleton which, by construction, must contain the target.

[3]Whenever $\rho$ and $E$ are unambiguous, we simply write $d_y$ rather than $d_y(\rho, E)$.
[4]*I.e.*, a set to which no more objects can be added.

A question in the above setup is how to select $\rho$: by Lemma 3, small values lead to a sharp decrease in the mass of Voronoi balls from one level to the next, hence reaching the target with fewer iterations. On the other hand, by Lemma 2, small values also imply larger nets, leading to more queries to the oracle per iteration. We select $\rho$ in an iterative fashion, as indicated in the pseudocode of Algorithm 2: we repeatedly halve $\rho$ until all non-singleton Voronoi balls $B_y(r_y)$ of the resulting net have a mass bounded by $0.5\mu(E)$. This selection leads to the following bounds on the corresponding query and computational complexity of RankNetSearch:

**Theorem 2.** RankNetSearch *locates the target by making* $4c^6(1+H(\mu))$ *queries to a comparison oracle, in expectation. The cost of determining which query to submit next is* $O\big(n(\log n + c^6)\log c\big)$.

In light of the $\Omega(cH(\mu))$ lower bound on query complexity by Karbasi *et al.* (2011), RankNetSearch is within a $O(c^5)$ factor of the optimal algorithm, and is thus order-optimal for constant $c$. Moreover, the computational complexity per query is $O\big(n(\log n + c^6\big)$, in contrast to the cubic cost of GBS. As shown in Section 6, in practice, this leads to drastic reductions in the computational costs compared to GBS.

The computational complexity can be further reduced to $O(1)$ through amortization. In particular, it is easy to see that the possible paths followed by RankNetSearch define a hierarchy, whereby every object serves as a parent to the rank net of its Voronoi ball. This tree can be pre-constructed, and search reduces descending this tree; we elaborate on this in Section 6.

## 5. Noisy Comparison Oracle

In a noisy setting the search must be robust against erroneous answers. Specifically, assume that for any query $\mathcal{O}_t(x,y)$, the *noisy oracle* returns the wrong answer with probability bounded by $\epsilon$, for some $\epsilon < 1/2$, *independently of previous answers*. In this context, a problem with RankNetSearch arises in line 4 of Algorithm 1: it is not clear how to identify the object closest to the target among elements in a net. We resolve this by introducing repetitions at each iteration. Specifically, at the $\ell$-th step of the search, $\ell \geq 1$, and rank-net size $m$, we define a repetition factor

$$k_\delta(\ell, m) := \frac{2\log\big((\ell+1/\delta)^2\lceil\log_2(m)\rceil\big)}{(1-\epsilon)^2} \tag{5}$$

for some design parameter $\delta \in (0,1)$. The modified algorithm then proceeds down the hierarchy, starting at the top level for $\ell = 1$. The basic step at step $\ell$ with a net $\mathcal{R}$ proceeds as follows. A *tournament* is organized among elements of $\mathcal{R}$, who are initially paired. Pairs of competing members are compared $k_\delta(\ell, |\mathcal{R}|)$ times. The "player" from a given pair winning the largest number of games moves to the next stage, where it will be paired again with another winner of the first round, and so forth until only one player is left. Note that the number of repetitions $k_\delta(\ell, m)$ increases only logarithmically with the level $\ell$.

To find the closest object to target $t$ with the noiseless oracle, clearly we need to make $O(|\mathcal{R}|)$ number of queries. The proposed algorithm achieves the same goal with high probability by making at most a factor $2k_\delta(\ell, |\mathcal{R}|)$ more comparisons. In this context we have the following

**Theorem 3.** *For a comparison oracle with error probability* $\epsilon$*, the algorithm with repetitions (5) outputs the correct target with probability at least* $1-\delta$ *in* $O(\frac{1}{(\frac{1}{2}-\epsilon)^2}\sum_{x\in\mathcal{N}}\mu(x)\log\frac{1}{\mu(x)}\log(\frac{1}{\delta}+\log\frac{1}{\mu(x)}))$ *queries, with constants depending on* $c$.

The proof is given in Appendix E. For uniform distribution $\mu(x) \equiv 1/n$, for all $x \in \mathcal{N}$, this yields an extra $\log\log(n)$ factor in addition to the term of order $H(\mu) = \log(n)$ which, by the lower bound by Karbasi *et al.*, is optimal.

## 6. Numerical Evaluation

We evaluate RankNetSearch over six publicly available datasets: *iris*, *abalone*, *ad*, *faces*, *swiss roll*, and *netflix*. We subsampled the latter two, taking 1000 randomly selected data points from *swiss roll*, and the 1000 most rated movies in *netflix*. We map these datasets to $\mathbb{R}^d$ (categorical variables are mapped to binary values in the standard fashion) for $d$ as shown in Fig. 1(a). For *netflix*, movies were mapped to 50-dimensional vectors by obtaining a low rank approximation of the user/movie rating matrix through SVD. For all experiments, the distance metric $d$ is the $\ell_2$ distance and the prior $\mu$ is power-law with $\alpha = 0.4$.

We evaluated the performance of two versions of RankNetSearch: one as described by Algo. 1, and another one (T-RankNetSearch) in which the hierarchy of rank nets is precomputed and stored as a tree. Both propose exactly the same queries to the oracle, so have the same query complexity; however, T-RankNetSearch has only $O(1)$ computational cost per query. The sizes of the trees precomputed by T-RankNetSearch for each dataset are shown in Fig. 1(a).

We compare these algorithms to (a) the policy proposed by Karbasi *et al.* (2011), denoted by Memo-

(a) Dataset Description

| Dataset | Size | Dim. | Tree |
|---|---|---|---|
| *iris* | 147 | 7 | 399 |
| *abalone* | 4177 | 11 | 21707 |
| *faces* | 698 | 4096 | 3664 |
| *ad* | 1924 | 1559 | 5816 |
| *swiss roll* | 1000 | 3 | 4229 |
| *netflix* | 1001 | 50 | 14585 |

Sources: `www.netflixprize.com`, `http://isomap.stanford.edu/datasets.html`, `http://archive.ics.uci.edu/ml`

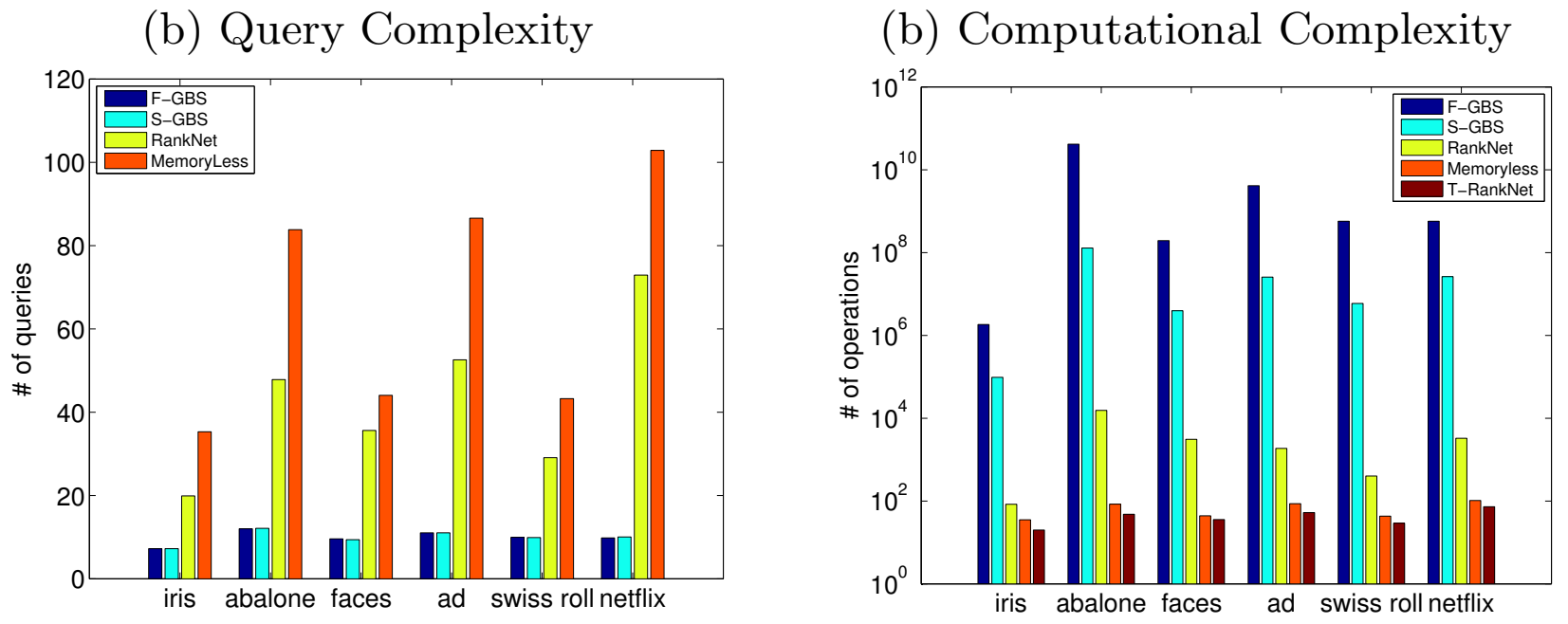

*Figure 1.* (a) Table of size, dimension (number of features), as well as the size of the Rank Net Tree hierarchy constructed for each dataset. (b) Expected query complexity, per search, of the five algorithms applied on each data set. As RankNet and T-RankNet have the same query complexity, only one is shown. (c) Expected computational complexity, per search, of the five algorithms applied on each dataset. For Memoryless and T-RankNet this equals the query complexity.

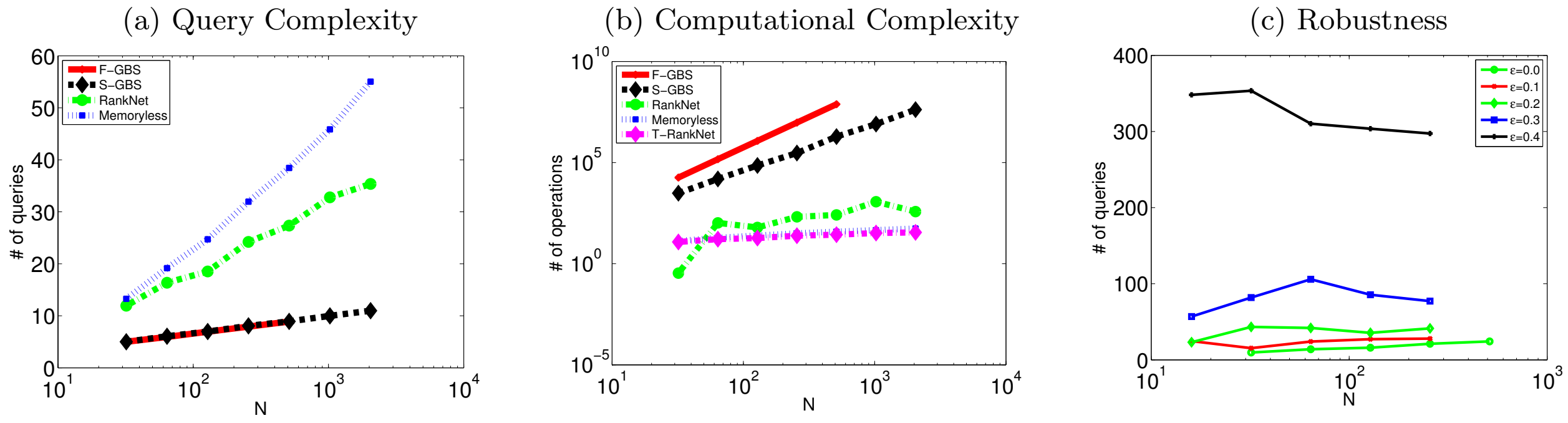

*Figure 2.* (a) Query and (b) computational complexity of the five algorithms as a function of the dataset size. The dataset is selected *u.a.r.* from the $\ell_1$ ball of radius 1.(c) Query complexity as a function of $n$ under a faulty oracle.

ryless, and (b) two heuristics based on GBS (the $\Theta(n^3)$ computational cost of GBS per query makes it intractable over the datasets we consider). The first heuristic, termed F-GBS for *fast* GBS, selects like GBS the query that minimizes (2); however, it does so by restricting the queries to pairs of objects in the current version space $V$. This reduces the computational cost per query to $\Theta(|V|^3)$, rather than $\Theta(n^2|V|)$. The second heuristic, termed S-GBS for *sparse* GBS, exploits rank nets as follows. First, we costruct the rank-net hierarchy over the dataset, as in T-RankNetSeach. Then, we minimize (2) restricted *only* on pairs of objects that appear in the same net. Intuitively, S-GBS assumes that an equitable partition of the objects exists among such pairs.

**Query vs. Computational Complexity.** The query complexity of different algorithms, expressed as average number of queries per search, is shown in Fig. 1(b). Although there are no known guarantees for either F-GBS nor S-GBS both algorithms are excellent in terms of query complexity across all datasets, finding the target within about 10 queries, in expectation. As GBS should perform as well as these algorithms, these suggest that it should also have low query cost. The query complexity of RankNet-Search is between 2 to 10 times higher; the impact is greater for high-dimensional datasets, as expected through the dependence of the rank net size on the $c$ doubling constant. Finally, Memoryless performs worse compared to all other algorithms. As shown in Fig. 1, the above ordering is fully reversed w.r.t. computational costs, measured as the aggregate number of operations performed per search. Differences from one algorithm to the next range between 50 to 100 orders of magnitude. F-GBS requires close to $10^9$ operations in expectation for some datasets; in contrast, RankNetSearch ranges between 100 and 1000 operations and, in conclusion, presents an excellent trade-off between query and computational complexity.

**Scalability and Robustness.** To study how the above algorithms scale with the dataset size, we also evaluate them on a synthetic dataset comprising objects placed uniformly at random at $\mathbb{R}^3$. The query and computational complexity of the five algorithms is shown in Fig. 2(a) and (b).

We observe the same discrepancies betwen algorithms we noted in Fig. 1. The linear growth in terms of $\log n$ implies a linear relationship between both measures of complexity w.r.t. the entropy $H(\mu)$ for all methods (we ommit the relevant figure for lack of space). In Fig. 2(b), we plot the query complexity of the robust RankNetSearch algorithm outlined in Section 5. For all simulations, the target success rate was set to 0.9, but the actual success rates we observed were considerably higher, close to 0.99. We observe that, even for high error rates $\epsilon$, the query complexity remains low. Moreover, the high success rates that we observe, combined with the independence of the cost on $n$, suggest that we can further reduce the number of queries to lower values than the ones required by (5).

# 7. Conclusion

We presented RankNetSearch, an algorithm that strikes an excellent balance between query and computational costs. Further improving this trade-off, in particular for more general kinds of noise, is an interesting future direction for this line of work. Throughout, we assumed that human inference of proximity is accurately captured by a metric space structure. An interesting research direction is assessing the validity of this assumption through user trials.

# A. Proof of Lemma 1

Using the ordered list containing the sets of equidistant objects described in Section 3.1, for any $z \in \mathcal{N}$, we can partition $\mathcal{N}$ into equivalence classes $A_1^z, A_2^z, \ldots, A_k^z$ such that for any two objects $y, y' \in \mathcal{N}$, $y \in A_i^z$ and $y' \in A_j^z$ with $i < j$ if and only if $d(y,z) < d(y',z)$.

To construct $\mathcal{R}$, it suffices to show that (4) can be verified for any $z, z' \in E$ using only the above partition and $\mu$. If so, a $\rho$-rank net can be constructed in a greedy fashion as a maximal set whose points verify (4). This can be obtained by adding sequentially an arbitrary object to the net and excluding from future selections any nodes that violate (4) w.r.t. this newly added object. Indeed, for all $y \in E$, $B_y(d_y) = \bigcup_{j=1}^{\ell} A_j^y$, where $\ell = \inf\{i : \sum_{j=1}^{i} \mu(A_j^y) \geq \rho\mu(E)\}$. The statement thus follows as (4) is equivalent to $y' \notin B_y(d_y) \vee y \notin B_{y'}(d_{y'})$. To construct the Voronoi balls $B_y(r_y) \subseteq E$, $y \in \mathcal{R}$, we initialize each such ball to contain its center $y$. For each $z \in E \setminus \mathcal{R}$, let $j_{\min}$ be the smallest $j$ such that $\mathcal{R} \cap A_j^z \neq \emptyset$; the object $z$ is then added to the ball $B_y(r_y)$ of every $y \in \mathcal{N} \cap A_{j_{\min}}^z$.

For each $y$, $B_y(d_y)$ can constructed in $O(\log|E|)$ time via binary search on the ordered list of equidistant objects. Constructing the rank net in a greedy fashion requires determining which objects violate (4) w.r.t. a newly added object on the net, which may take $O(|E|)$ time. Hence, the overall complexity of constructing $\mathcal{R}$ is $O(|E|(|\mathcal{R}| + \log|E|))$. Finally, the construction of the Voronoi balls requires $O(|\mathcal{R}|)$ steps per object in $E$ to assign each object to a ball. □

# B. Proof of Lemma 2

Note first that, for all distinct $y, y' \in \mathcal{R}$, the balls $B(y, d_y(\rho, E)/4) \cap B(y', d_{y'}(\rho, E)/4) = \emptyset$. To see this, assume w.l.o.g. that $d_y \geq d_{y'}$ which implies that $d(y, y') \geq d_y - d_{y'}$. This is due to the fact that $\mu(B(y, d_{y'})) \geq \rho\mu(E)$, and hence, by (3), $d_y$ can be at most $d_{y'} + d(y, y')$. In case $d_y$ or $d_{y'}$ is zero, clearly $d(y,y') > d_y/2 > d_y/4 + d_{y'}/4$. If $0 < d_{y'} < d_y/2$, then $d(y,y') > d_y/2 \geq d_y/4 + d_{y'}/4$.. If $d_{y'} \geq d_y/2 > 0$, then $d(y,y') \geq d_{y'} \geq d_y/2 > d_y/4 + d_{y'}/4$.. Hence, in all cases $d(y,y') > d_y/4 + d_{y'}/4$ and as a result $B(y, d_y/4) \cap B(y', d_{y'}/4) = \emptyset$.

To prove Lemma 2, observe that $d_y \leq 2R$ for all $y \in \mathcal{R}$ since $\mu(y,d) \geq \mu(E) > \rho\mu(E)$ for $d \leq 2R$. Therefore, $d_y/4 \leq R/2$ and thus $B(y, d_y/4) \subseteq B(x, 2R)$. Hence, by the definition of $c(\mu)$, $\sum_{y\in\mathcal{R}} \mu(B(y, d_y/4)) \leq \mu(B(x, 2R)) \leq c\mu(E)$. Moreover, $\sum_{y\in\mathcal{R}} \mu(B(y, d_y/4)) \geq c^{-2} \sum_{y\in\mathcal{R}} \mu(B(y, d_y)) \geq c^{-2}\rho\mu(E)|\mathcal{R}|$. Therefore, $|\mathcal{R}| \leq c^3/\rho$. □

## C. Proof of Lemma 3

Observe first that, for all $z \in E$, there exists a $y \in \mathcal{R}$ such that $z \in B(y, d_y(\rho, E))$. To see this, assume otherwise. Then for any $y \in \mathcal{R}$, $d(z,y) > d_y(\rho,E) \geq \min\{d_y(\rho,E), d_z(\rho,E)\}$ and we can add $z$ to $\mathcal{R}$, which contradicts its maximality.

To prove Lemma 3, we consider the following two cases. Suppose first $0 < r_y \leq d_y$. By (3), for any $\tilde{r} < d_y$, we have $\mu(B(y,\tilde{r}) < \rho\mu(E)$. In particular, $\mu(B(y, d_y/2) < \rho\mu(E)$. By the definition of $c$, $\mu(B(y, r_y) \leq \mu(B(y, d_y) \leq c\rho\mu(E)$. For the second case, suppose that $r_y > d_y$. Let $z \in V_y$ is the point for which $d(y,z) = r_y$. By the above observation, we know that there exists a $y' \in \mathcal{R}$ such that $d(z,y') \leq d_{y'}$. As $r_y > d_y$, $y \neq y'$. On the other hand, $d(z,y') \geq d(z,y)$ since $z \in V_y$. Using the triangle inequality, we get $d(y,y') \leq d(y,z) + d(y',z) \leq 2d(y',z) \leq 2d_{y'}$. We know that $B(y, r_y) \subseteq B(y', d(y,y') + r_y)$. Since $r_y = d(y,z) \leq d_{y'}$ we can say $B(y, r_y) \subseteq B(y', 3d_{y'})$. Finally, by the definition of $c$, we have $\mu(B(y, r_y)) \leq \mu(B(y', 3d_{y'})) \leq c^2\mu(B(y', d_{y'})) \leq c^3\rho\mu(E)$. □

## D. Proof of Theorem 2

Note first that, by induction, it can be shown that the version space is a subset of $E$; correctness is implied by this fact and the termination condition. To bound the number of queries, we first show that the process RankNet constructs a net with small cardinality.

**Lemma 4.** RankNet *terminates at* $\rho > \frac{1}{4c^3}$.

*Proof.* To see that the **while** loop terminates, observe that, by Lemma 3, for small enough $\rho < \min_{z\in E} \mu(z)/(c^3\mu(E))$, all Voronoi balls $B_y(r_y)$ of the $\rho$-rank net $\mathcal{R}$ will be singletons, so $\mathcal{I}$ will indeed be empty. Suppose thus that the loop terminates at some $\rho = \rho^*$. Since it did not terminate at $\rho = 2\rho^*$, there exists a ball $B_y(r_y)$ of the $2\rho$-rank net $\mathcal{R}$ such that $r_y > 0$ and $\mu(B_y(r_y)) > 0.5\mu(E)$. By Lemma 3, $\mu(B_y(r_y)) \leq c^3 2\rho\mu(E)$, and the lemma follows. □

Hence, from Lemmas 2 and 4, we get that the rank nets returned by RankNet have cardinality at most $4c^6$. On the other hand, by construction, a net covering a ball $B_y r_y$ consists of either singletons or balls with mass less than $0.5\mu(B_y(r_y))$. As a result, at each iteration, moving to the next object either halves the mass of the version space or leads to a leaf, and the search terminates. As at any point the version space has a mass greater than $\mu(t)$, the search will terminate after traversing most $\lceil \log_2(1/\mu(t)) \rceil$ iterations. Since, at each level, the number of accesses to the oracle are $\mathcal{R} - 1 \leq 4c^3$, the total query cost for finding target $t$ is at most $4c^6 \lceil \log_2(1/\mu(t)) \rceil$, and the query complexity statement follows. Finally, from Lemmas 1 and 4, the computational complexity of each RankNet call is at most $O\big(n(\log n + c^3)\log c\big)$. □

## E. Proof of Theorem 3

We first show the following auxiliary result.

**Lemma 5.** *Given a target $t$ and a noisy oracle with error probability bounded by $\epsilon$, the tournament among elements of the net $\mathcal{R}$ with repetitions $k_\delta(\ell, |\mathcal{R}|)$ returns the element in the set $\mathcal{R}$ that is closest to target $t$ with probability at least* $1 - (\ell + 1/\delta)^{-2}$.

*Proof.* We assume for simplicity that there are no ties, i.e. there is a unique point in $\mathcal{R}$ that is closest to $t$. The case with ties can be deduced similarly. We first bound the probability $p(k)$ that upon repeating $k$ times queries $\mathcal{O}_t(x,y)$, among $x$ and $y$ the one that wins the majority of comparisons is not the closest to $t$. Because of the bound $\epsilon$ on the error probability, one has $p(k) \leq \Pr(\text{Bin}(k,\epsilon) \geq k/2)$, where $\text{Bin}(\cdot,\cdot)$ denotes the Binomial distribution. Azuma-Hoeffding inequality ensures that the right-hand side of the above is no larger than $\exp(-k(1/2-\epsilon)^2/2)$. Upon replacing the number of repetitions $k$ by the expression (5), one finds that $p(k_\delta(\ell, |\mathcal{R}|)) \leq (\ell + 1/\delta)^{-2} / \lceil \log_2(|\mathcal{R}|) \rceil$. Consider now the games to be played by the element within $\mathcal{R}$ that is closest to $t$. There are at most $\lceil \log_2(|\mathcal{R}|) \rceil$ such games. By the union bound, the probability that the closest element loses on any one of these games is no less than $(\ell + 1/\delta)^{-2}$, as announced. □

By the union bound and the previous Lemma we have conditionally on any target $t \in \mathcal{N}$ that $\Pr(\text{success}|T = t) \geq 1 - \sum_{\ell\geq 1}(\ell + 1/\delta)^{-2})$. The latter sum is readily bounded by $\delta$. The number of comparisons given that the target is $T = t$ is at most $\sum_{\ell=1}^{\lceil \log_2(1/\mu(t)) \rceil} 2|\mathcal{R}_\ell| k_\delta(\ell, |\mathcal{R}_\ell|) = O\left(\frac{1}{(\frac{1}{2}-\epsilon)^2} \log\frac{1}{\mu(t)} \log(\frac{1}{\delta} + \log\frac{1}{\mu(t)})\right)$, where the $O$-term depends only on the doubling constant $c$. The bound on the expected number of queries follows by averaging over $t \in \mathcal{N}$. □